\documentclass[conference]{IEEEtran}
\IEEEoverridecommandlockouts

\usepackage{cite}
\usepackage{amsmath,amssymb,amsfonts}
\usepackage{algorithm}
\usepackage{algorithmic}
\usepackage{graphicx}
\usepackage{textcomp}
\usepackage{xcolor}
\usepackage{todonotes}
\usepackage{enumitem}
\usepackage{booktabs}
\usepackage{subfigure}
\usepackage{balance}
\usepackage{url}
\usepackage{csquotes} 
\usepackage{tikz}     
\usetikzlibrary{arrows.meta,positioning,fit,backgrounds,calc,shapes.geometric}
\tikzset{
  phx/.style={
    font=\footnotesize,
    box/.style={rounded corners=1pt, draw=black!70, line width=0.5pt,
                align=center, inner sep=2pt},
    host/.style={box, fill=black!10, minimum height=5mm, minimum width=16mm},
    comp/.style={box, fill=blue!8, align=center},
    sub/.style={box, fill=blue!55!black, text=white, minimum height=4.2mm,
                minimum width=17mm, inner xsep=1pt, font=\scriptsize},
    stage/.style={box, fill=blue!8, minimum height=5mm, font=\scriptsize,
                 inner xsep=2pt},
    art/.style={box, fill=orange!12, font=\scriptsize, inner xsep=2pt,
                minimum height=5mm},
    slot/.style={box, fill=orange!70!black, text=white, minimum height=4.2mm,
                 inner xsep=2pt, font=\scriptsize},
    choice/.style={box, fill=yellow!15, font=\scriptsize, align=center,
                inner xsep=2pt, minimum height=5mm},
    good/.style={box, fill=green!12, font=\scriptsize, inner xsep=2pt,
                 minimum height=5mm},
    bad/.style={box, fill=red!10, font=\scriptsize, inner xsep=2pt,
                minimum height=5mm},
    band/.style={box, inner sep=3pt},
    title/.style={font=\footnotesize\bfseries, inner sep=1pt, anchor=west},
    ctitle/.style={font=\footnotesize\bfseries, inner sep=1pt},
    flow/.style={-{Latex[length=1.4mm]}, line width=0.5pt},
    lbl/.style={inner sep=1pt, fill=white, font=\scriptsize},
    graph/.style={box, fill=violet!12, draw=violet!60, font=\scriptsize,
                 inner xsep=2pt, minimum height=5mm, align=left},
    note/.style={font=\scriptsize, inner sep=1pt, text=black!65},
  },
}

\usepackage{pifont}   

\definecolor{okgreen}{HTML}{0B5A2B}  
\definecolor{badred}{HTML}{B3261E}   
\usepackage[nolist,nohyperlinks]{acronym}
\begin{acronym}
  \acro{MCP}{Model Context Protocol}
  \acro{LLM}{large language model}
\end{acronym}
\def\BibTeX{{\rm B\kern-.05em{\sc i\kern-.025em b}\kern-.08em
    T\kern-.1667em\lower.7ex\hbox{E}\kern-.125emX}}

\def\phronix{Phronix}
\def\cluster{CLUSTER}

\begin{document}

\title{\phronix{}: The Capability-Centric Learning Paradigm for Agents}
\author{\IEEEauthorblockN{Hariharan Devarajan$^1$, Neeraj Rajesh$^2$, Aniruddha Marathe$^1$, Nikoli Dryden$^1$, Kin Ng$^3$, and Michela Taufer$^3$}
\IEEEauthorblockA{$^1$Lawrence Livermore National Laboratory, $^2$Illinois Institute of Technology, and $^3$University of Tennessee, Knoxville \\
hariharandev1@llnl.gov, nrajesh@hawk.iit.edu, marathe1@llnl.gov, dryden1@llnl.gov, kng2@utk.edu, and taufer@utk.edu}}
\pagestyle{plain}
\maketitle

\begin{abstract}
Scientific computing increasingly relies on large language model-based agents for performance engineering tasks such as instrumentation, analysis, and workflow porting.
However, agent effectiveness depends heavily on the surrounding harness, which typically relies on manually written skills and curated memory. 
This limits autonomous learning, forcing agents to repeatedly rediscover solutions rather than retain and reuse knowledge from prior executions.
We introduce \phronix{}, a capability-centric learning paradigm that enables agentic harnesses to acquire, validate, and refine reusable capabilities from its own executions.  \phronix{} operates through a predictor--validator--corrector loop: MCP tools and associated skills propose solutions, tool executions validate them, and observed outcomes refine the stored capabilities. 
It distills every observed interaction into a capability store and researches a failing task rather than abandoning it.
First, such learned capabilities solve a recurring task on the first attempt in 93--95\% of cases across every model, significantly reducing solution non-determinism and reliability.
Second, \phronix{} turns an agent's own execution into 7.6--30.8 durable capabilities per task without user instruction, and creation falls below two once the needed capability exists.
Third, a warm-start runs consume only  0.38--0.40 million tokens per task which is three to four times less than a single baseline attempt, while a cold-start costs no more than the baseline.
On the 1000 Genome Pegasus workflow, instrumenting a production application for I/O performance analysis, \phronix{} passes all 11 correctness checks against six for the baseline harness, and also lifts the task solving rates of two cheaper models from 29\% and 27\% of SWE-bench tasks solved to 91\% and 87\%, respectively, matching the performance of the most expensive evaluated model.\looseness=-1
\end{abstract}


\section{Introduction}
\label{sec:introduction}

Scientific computing has begun to delegate its routine engineering to \ac{LLM} agents, which auto-tune HPC code on production supercomputers~\cite{hayashi_vibecodehpc_2026}, diagnose parallel I/O performance~\cite{egersdoerfer_ioagent_2025}, and work on the repositories and pipelines behind published results~\cite{duston_ainsteinbench_2025,bragg_astabench_2026}.
The work delegated there is repetitive rather than novel, such as instrumenting an application, characterizing where a run spends its time, and repeating that for the next application, input scale, and system.
An \emph{agent}, however, is more than a model which operates withing a \emph{harness}~\cite{xi_rise_2023}, where the \ac{LLM} proposes each step and observes its result~\cite{yao_react_2023}, \emph{tools} exposed through interfaces such as the \ac{MCP}~\cite{schick_toolformer_2023,anthropic_mcp_2024} let it touch the world, \emph{skills} accumulate how a class of task is done~\cite{wang_voyager_2023,anthropic_claude_code}, \emph{memory} carries what earlier sessions learned past the context window~\cite{packer_memgpt_2023}, and the harness orchestrates them into the loop the agent runs.
An agent that is good at such a task once should be good at it every time, and it is the harness, not the model, that determines whether it is, because the harness governs how much human steering a task requires and how many tokens and retries it consumes.
Improving agent effectiveness is consequently as much a harness problem as it is a model problem.\looseness=-1


Traditionally, \acp{LLM} were accessed through web chat interfaces such as Open WebUI~\cite{open_webui}, ChatGPT~\cite{openai_chatgpt}, and Claude~\cite{anthropic_claude}, which wrap a model in a conversation backed by a fixed set of tools~\cite{nakano_webgpt_2021,schick_toolformer_2023,qin_toolllm_2024}.
As these interfaces matured, developers built more complex and capable command-line harnesses such as Codex~\cite{openai_codex_cli} and OpenCode~\cite{opencode}, which let users plug in different skills and agents for each task.
State-of-the-art harnesses such as Claude Code~\cite{anthropic_claude_code} and Stakpak~\cite{stakpak} add memory that persists across sessions, together with rulebooks that curate the standard operating procedure for a class of task.
Along this timeline, each harness has become more adaptable than the last, and that adaptation is increasingly carried by skills and memory rather than by the interface itself~\cite{wang_voyager_2023,shinn_reflexion_2023,park_generative_2023,packer_memgpt_2023}.
All of them are driven primarily by human, who writes the skill or rulebook that a complex task needs before it is assigned.\looseness=-1

Despite support for skills and persistent memory, existing harnesses often rely on users to decide what knowledge to retain and reuse, limiting their ability to learn autonomously from completed tasks.
For an agent to learn from its own work instead, the harness must overcome three challenges that each keep the user in the learning loop.
First, \emph{solution non-determinism} prevents an agent from reproducing a solution it has already found, because the skills and models of the harness keep changing how they solve the same problem, so a task solved once may be solved differently or not at all when it recurs.
Second, \emph{manual tracking} requires users to identify and preserve useful information from agent executions, so the experience a task produced is lost before it can be reused.
Third, \emph{manual learning} leaves the conversion of that experience into reusable knowledge to a user who must ask for a skill by name, so an agent that repeats a task never becomes better at it.
These challenges are particularly significant in scientific computing, where the same instrumentation, performance analysis, and workflow porting recur across applications and systems.
Addressing them requires a learning paradigm in which agents organically build, correct, and learn capabilities without waiting for a user to do so on their behalf.\looseness=-1

We present \phronix{}, a learning paradigm that makes building capabilities the first-class citizen of the harness and runs every one of them as a predictor--validator--corrector loop, where a capability generates an artifact, tool-based validation deterministically rules on its correctness, and the findings of that ruling correct the capability for future use rather than working around it.
\phronix{} addresses the three challenges through complementary mechanisms.
First, \emph{deterministic but intelligent capabilities} combine an \ac{MCP} tool for reproducible execution with a skill that provides contextual judgment and handles corner cases, reducing solution non-determinism.
Second, \emph{automatic tracking and knowledge creation}, captures the inputs, execution steps, and outputs of harness--model interactions and distills them into reusable knowledge without user intervention. 
Third, \emph{continual-learning-based research}, investigates failed tasks, incorporates findings into the capability store, and adapts retry effort based on the progress toward a solution.
To ensure reliability, a validator admits a capability only if it compiles and is not a duplicate, groups a distilled observation arithmetically rather than asking the model what it meant, and marks a researched approach working only by measurement on held-out instances, while the improvement a task shows between attempts is the convergence criterion that decides whether it continues.
Together, these three concepts make \phronix{} self-learning because it continuously builds, corrects, and reuses capabilities without a user in the loop. 
This paper makes the following contributions:\looseness=-1

\begin{enumerate}[leftmargin=1.3em]
  \item We introduce a \emph{capability-centric learning paradigm}  that enables agentic harnesses to autonomously build, validate, refine, and reuse executable capabilities through a predictor--validator--corrector loop, replacing reliance on user-curated memory with learning from execution.
  \item We design deterministic and intelligent capabilities that pair an \ac{MCP} tool with the skills that provide the contextual judgment and handles corner cases that cannot be coded.
  \item We design \emph{automatic tracking} that records the input, execution, and output of every harness--model interaction and distills them into skills and \ac{MCP} tools without user instruction.
  \item We design \emph{continual-learning-based research} that turns a failing task into research whose outcome, solved or failed, feeds back into the capability store, with an attempt budget that follows measured progress.
  \item We evaluate \phronix{} on SWE-bench and on an HPC use case, instrumenting the 1000 Genome Pegasus workflow~\cite{deelman_pegasus_2015,1000genomes_global_2015} with DFTracer~\cite{devarajan_dftracer_2024} for I/O performance analysis, showing that a warm-start run costs 3.2--3.9$\times$ fewer tokens per task than one baseline attempt while lifting the solve rate from 27--65\% to 87--91\%.
\end{enumerate}

\section{Background}
\label{sec:background}

Agents built on \acp{LLM} have entered the scientific-computing toolchain, auto-tuning HPC code on production supercomputers~\cite{hayashi_vibecodehpc_2026}, diagnosing parallel I/O performance~\cite{egersdoerfer_ioagent_2025}, and operating on scientific repositories and research pipelines~\cite{duston_ainsteinbench_2025,bragg_astabench_2026}, where the tasks they are given recur across applications, input scales, and machines.
What such an agent retains from one instance therefore depends on the architecture of the agentic system built around the model and the learning mechanisms it offers today.\looseness=-1

\subsection{Agentic System Architecture}
\label{sec:background-agentic}

An \ac{LLM} generates responses from a given prompt, enabling tasks such as writing code, reading a stack trace, and planning a multi-step change~\cite{jimenez_swe-bench_2024}. Since models vary considerably in capability and cost, practitioners often choose models based on task complexity and requirements.
A model is nevertheless stateless and stochastic, sampling its answer afresh on every call and keeping nothing beyond what the caller puts back into a bounded prompt, so capability arrives from the model while reliability and memory come from the loop around it.
That loop is what makes an agent, the model proposing the next step, the components around it carrying that step out, and the result becoming its next input.\looseness=-1

Because the model can only emit text, every step that touches the world is taken by a \emph{tool}, a callable unit of code such as a file read, a job submission, or a trace query, which the \ac{MCP} has made a common interface for exposing to any model~\cite{anthropic_mcp_2024}.
A tool does the same thing each time it is called but cannot decide when to be called, and a \emph{skill} shapes that decision, being a document loaded into context when its description matches the task at hand~\cite{anthropic_claude_code}.
It states how this class of task is done here, which tools to use in what order, and what to check before moving on, so a skill carries judgment rather than execution and is only as good as what its author knew.
\emph{Memory} spans the gap between sessions, holding user preferences and notes on past decisions that are reloaded on the next run, so what one session established is not derived again from scratch.\looseness=-1

The \emph{harness} coordinates these components through an interative execution loop~\cite{anthropic_claude_code,openai_codex_cli}.
At each step it selects which skills and memories enter a context window it must keep within budget, presents the tools the model may call, executes the call it returns, enforces provided execution permissions, and appends the result before asking for the next step.
The model therefore supplies capability, tools action, skills procedure, and memory continuity, while the harness decides which the model sees at a given step.
What it puts in front of the model is the whole of what the agent can do there, and anything it does not surface is work the agent must reconstruct from scratch.\looseness=-1

\subsection{Current Learning Capabilities}
\label{sec:background-learning}

Existing agentic harnesses retain knowledge primarily through \emph{memory}, 
which preserves preferences and past decisions across sessions, and \emph{skills}, 
which encode reusable task-specific procedures~\cite{anthropic_claude_code}. 
Other approaches include retrieval-augmented generation, which retrieves relevant documents at query time~\cite{lewis_retrieval-augmented_2020}, and fine-tuning, which adapts model parameters through a separate training process~\cite{hu_lora_2021}.
While these mechanisms support knowledge retention and reuse, 
they do not inherently capture successful executions as validated,
executable capabilities. 
Consequently, 
harnesses often depend on user-authored or curated knowledge, 
leaving agents to rediscover solutions rather than systematically learn from their own executions.\looseness=-1

\section{Motivation}
\label{sec:motivation}

Consider instrumenting the 1000 Genomes workflow~\cite{1000genomes_global_2015} with DFTracer~\cite{devarajan_dftracer_2024} to trace its I/O,  a task that requires an agent to install the tracer, annotate the read and per-individual write paths, and run the annotated application.
We gave this task to the Claude Code harness driving gpt-5.5, and the run finished with a working trace, yet exhibited all three challenges that keep such a harness from learning from its own work.\looseness=-1

\emph{Solution non-determinism.}
The agent solved the task by issuing 39 shell commands, 31 of which chained several steps using \texttt{\&\&} and the longest was 598 characters, none saved as a script.
They cannot be replayed as written because 7 embed the timestamped run directory created for that session, 13 read files under a session-specific temporary path, 3 poll for a task-output file named after a session identifier, and 4 load or purge modules whose effects depend on what the shell had loaded before.
The route was improvised as well, since the agent tried three installation routes before one succeeded, rebuilt the environment under a different interpreter, and swapped the compiler toolchain rather than adding the link flag it had itself diagnosed.
A later run therefore starts from nothing reusable, as the only record of the solution is commands written for one directory, one session, and one module state.\looseness=-1

\emph{Manual tracking.}
The run recorded nothing it learned, leaving no skill and no \ac{MCP} tool, because 49 of its 50 calls were shell-tool calls whose output existed only in a transcript discarded with the session.
Yet the transcript contained several details that could have been reused: the tracer needed an extra link flag, its build failed under the default compiler, and the workflow's dependencies were unnecessary for this input. None of this knowledge was written down.
A user can ask the harness to record such information, but must first know which of the 50 calls mattered can ask.\looseness=-1

\emph{Manual learning.}
Because nothing was distilled into an executable capability, a second run must rediscover the same information, repeating the 50 calls and processing the same 2.84M input tokens.
The agent also made poor decisions that reusable knowledge could have helped prevent. 
For example, it annotated the application's write loop but left the outer compression function without a span, resulting in 6 incorrect decisions out of the 11 decisions in the finished workflow.
An agent that cannot turn its execution into a capability keeps paying the full price of a task it has solved, while also risking the same mistakes because the knowledge from the previous run was not captured for reuse.\looseness=-1

These challenges arise from how the harness retains and reuses knowledge rather than of the model driving it, which is why we make building capabilities its first-class concern.\looseness=-1

\section{\phronix{}}
\label{sec:design}


\phronix{} makes an agent's knowledge durable, retrievable, and verifiable, so that knowledge established during one task can be reused in the next as something the agent can execute rather than read and interpret again.
Three objectives follow.
Knowledge must be \emph{actionable}, because prose is interpreted differently each time it is read, so anything with one right answer for given inputs belongs in code, while prose carries only the decisions that require judgment.
It must be \emph{acquired implicitly}, because requiring a user to first identify which of a hundred steps mattered would make the user the bottleneck the harness was meant to remove.
It must be \emph{verified}, because unverified knowledge becomes a liability when later tasks rely on it, so a model's prediction is admitted only after a deterministic validator has confirmed it and any required correction has been applied.\looseness=-1

\subsection{High-Level Design}
\label{sec:design-overview}

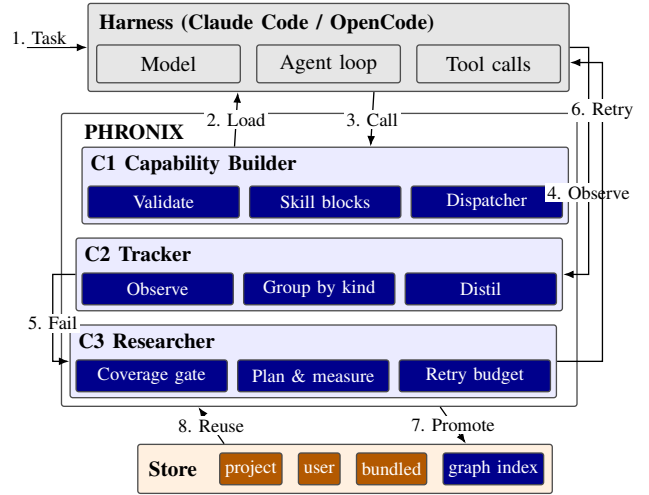
\begin{figure}[t]
\centering
%
\begin{tikzpicture}[
  font=\footnotesize,
  box/.style={rounded corners=1pt, draw=black!70, line width=0.5pt,
              align=center, inner sep=2pt},
  host/.style={box, fill=black!10, minimum height=5mm, minimum width=19mm},
  comp/.style={box, fill=blue!8, align=center},
  sub/.style={box, fill=blue!55!black, text=white, minimum height=4.2mm,
              minimum width=20mm, inner xsep=1pt, font=\scriptsize},
  slot/.style={box, fill=orange!70!black, text=white, minimum height=4.2mm,
               inner xsep=2pt, font=\scriptsize},
  band/.style={box, inner sep=3pt},
  title/.style={font=\footnotesize\bfseries, inner sep=1pt, anchor=west},
  ctitle/.style={font=\footnotesize\bfseries, inner sep=1pt},
  flow/.style={-{Latex[length=1.4mm]}, line width=0.5pt},
  lbl/.style={inner sep=1pt, fill=white, font=\scriptsize},
]

\node[host] (model) at (0,0) {Model};
\node[host, right=2mm of model] (loop) {Agent loop};
\node[host, right=2mm of loop] (tools) {Tool calls};
\node[title, above=1mm of model.north west, anchor=south west] (htitle)
  {Harness (Claude Code / OpenCode)};
\begin{scope}[on background layer]
  \node[band, fill=black!10, fit=(model)(loop)(tools)(htitle)] (harness) {};
\end{scope}

\node[ctitle, below=8mm of harness.south west, anchor=north west] (c1t)
  {C1~Capability Builder};
\node[sub, below=1mm of c1t.south west, anchor=north west] (c1a) {Validate};
\node[sub, right=1.2mm of c1a] (c1b) {Skill blocks};
\node[sub, right=1.2mm of c1b] (c1c) {Dispatcher};
\begin{scope}[on background layer]
  \node[comp, fit=(c1t)(c1a)(c1c)] (builder) {};
\end{scope}

\node[ctitle, below=2.5mm of builder.south west, anchor=north west] (c2t) {C2~Tracker};
\node[sub, below=1mm of c2t.south west, anchor=north west] (c2a) {Observe};
\node[sub, right=1.2mm of c2a] (c2b) {Group by kind};
\node[sub, right=1.2mm of c2b] (c2c) {Distil};
\begin{scope}[on background layer]
  \node[comp, fit=(c2t)(c2a)(c2c)] (tracker) {};
\end{scope}

\node[ctitle, below=2.5mm of tracker.south west, anchor=north west] (c3t) {C3~Researcher};
\node[sub, below=1mm of c3t.south west, anchor=north west] (c3a) {Coverage gate};
\node[sub, right=1.2mm of c3a] (c3b) {Plan \& measure};
\node[sub, right=1.2mm of c3b] (c3c) {Retry budget};
\begin{scope}[on background layer]
  \node[comp, fit=(c3t)(c3a)(c3c)] (research) {};
\end{scope}

\node[title, above=0.5mm of builder.north west, anchor=south west] (ptitle) {PHRONIX};
\begin{scope}[on background layer]
  \node[band, fill=none, fit=(builder)(tracker)(research)(ptitle)] (phronix) {};
\end{scope}

\node[slot, below=6mm of phronix.south, anchor=north] (user) {user};
\node[slot, left=2mm of user] (project) {project};
\node[slot, right=2mm of user] (bundled) {bundled};
\node[slot, right=2mm of bundled, fill=blue!55!black] (graph) {graph index};
\node[title, left=2.5mm of project, anchor=east] (stitle) {Store};
\begin{scope}[on background layer]
  \node[band, fill=orange!12, fit=(project)(user)(bundled)(graph)(stitle)] (store) {};
\end{scope}

\draw[flow] ([xshift=-8mm]harness.west) -- node[lbl,above,pos=0.2]{1.~Task} (harness.west);
\draw[flow] ([xshift=-12mm]builder.north) -- node[lbl,pos=0.5]{2.~Load}
            ([xshift=-12mm]harness.south);
\draw[flow] ([xshift=6mm]harness.south) -- node[lbl,pos=0.5]{3.~Call}
            ([xshift=6mm]builder.north);
\draw[flow] (harness.east) -- ++(2.5mm,0) |- node[lbl,pos=0.32]{4.~Observe} (tracker.east);
\draw[flow] (tracker.west) -- ++(-3mm,0) |- node[lbl,pos=0.28]{5.~Fail} (research.west);
\draw[flow] (research.east) -- ++(6mm,0) |- node[lbl,pos=0.42]{6.~Retry} ([yshift=-2mm]harness.east);
\draw[flow] ([xshift=16mm]phronix.south) -- node[lbl,pos=0.5]{7.~Promote}
            ([xshift=16mm]store.north);
\draw[flow] ([xshift=-16mm]store.north) -- node[lbl,pos=0.5]{8.~Reuse}
            ([xshift=-16mm]phronix.south);
\end{tikzpicture}
\vspace{-3.5mm}
\caption{\phronix{} adds three components to an existing harness, one per concept, over a capability store whose three layers travel with the framework, the user, and the repository. The numbered arrows follow one agentic task.}
\vspace{-4mm}
\label{fig:high-level}
\end{figure}

Figure~\ref{fig:high-level} attaches \phronix{} to a harness already running an agent loop over a model, and follows one task through the three components.
A task arrives as it would without \phronix{} (arrow~1), and the Capability Builder loads the skills and \ac{MCP} tools the store holds for such work into the model's context (arrow~2), so the agent begins with capabilities established by earlier tasks rather than with an empty environment.
The agent solves the task by calling those capabilities (arrow~3), and every interaction with the harness, whether successful or failed, reaches the Tracker as it happens (arrow~4).
When an attempt fails, the Researcher reads that record (arrow~5), decides whether the store already covers the problem, plans an approach if it does not, and returns the task with an attempt budget that follows measurable progress (arrow~6).
Whatever the task produced, the Tracker distills its observations into skills and \ac{MCP} tools, and promotes them (arrow~7), which the next task loads (arrow~8).\looseness=-1

All three components separate prediction, validation, and correction, so the model never judges its own output.
A capability \emph{predicts}, pairing an \ac{MCP} tool with the skills that carry the judgment around it.
A validator \emph{validates}, being a tool alone and therefore returning the same verdict every time, whether it compiles and fingerprints a new capability, groups an observation arithmetically, or marks an approach working only by measurement on held-out instances.
Its verdict names each fault in enough detail to repair without diagnosing it again, and the agent \emph{corrects} the capability or its wiring rather than working around it, which turns a failed check into a better capability instead of a one-off patch.\looseness=-1

Validators are written alongside the task that produces what they check, so prediction and validation evolve together, and \phronix{} applies them at two levels.
Structural validators are built-in to check every skill, capability and agent: frontmatter must parse, links must resolve, modules must compile and register their documented tools, and templates must render for every harness. This keeps a store's verified fraction from starting at zero.
Authored verifiers specify claims that structure alone cannot settle, declaring their assumptions, inputs, constraints, invariants, and outputs separately, so a machine missing a dependency produces an inconclusive result rather than a failure. Genuine failures therefore remain distinguishable, and each verifier travels with the capability it checks.
The loop converges on the improvement a task shows between attempts rather than on a model's confidence, so a task that improves earns further attempts, while repeated failures cause it to stop.\looseness=-1

The store is the substrate shared by the three components rather than a fourth component, and its three layers decide how far a capability travels, from the bundled layer through a user's layer to the project layer committed with the repository.
A knowledge graph makes it usable at scale, since every capability, skill block, and research record is a node that a task can locate by querying, allowing each component to retrieve only the objects it needs instead of paying the context cost for the rest.\looseness=-1

\subsection{Capability Builder: Deterministic and Intelligent}
\label{sec:design-c1}

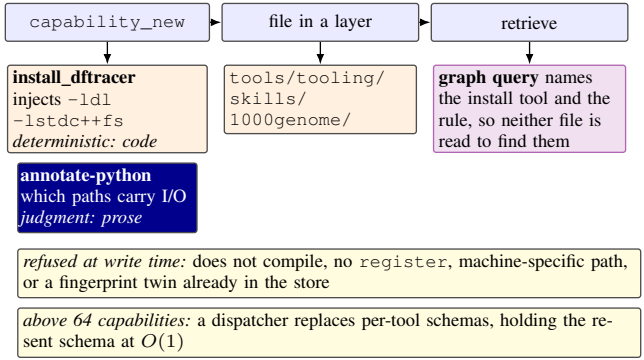
\begin{figure}[t]
\centering
\begin{tikzpicture}[phx]
\node[stage, minimum width=27mm] (write) {\texttt{capability\_new}};
\node[stage, right=1.5mm of write, minimum width=26mm] (file) {file in a layer};
\node[stage, right=1.5mm of file, minimum width=26mm] (get) {retrieve};

\node[art, below=3mm of write.south, anchor=north, align=left,
      text width=25mm] (w1)
  {\textbf{install\_dftracer}\\
   injects \texttt{-ldl -lstdc++fs}\\
   \emph{deterministic: code}};
\node[sub, below=1.2mm of w1.south, anchor=north, text width=23mm,
      align=left] (w2)
  {\textbf{annotate-python}\\ which paths carry I/O\\ \emph{judgment: prose}};
\node[art, below=3mm of file.south, anchor=north, align=left,
      text width=24mm] (f1)
  {\texttt{tools/tooling/}\\ \texttt{skills/}\\ \texttt{1000genome/}};
\node[graph, below=3mm of get.south, anchor=north,
      text width=24mm] (g1)
  {\textbf{graph query} names the install tool and the rule, so neither file is read to find them};

\draw[flow] (write) -- (file);
\draw[flow] (file) -- (get);
\draw[flow] (write.south) -- (w1.north);
\draw[flow] (file.south) -- (f1.north);
\draw[flow] (get.south) -- (g1.north);

\node[choice, below=2mm of w2.south west, anchor=north west, align=left,
      text width=81mm] (checks)
  {\emph{refused at write time:} does not compile, no \texttt{register},
   machine-specific path, or a fingerprint twin already in the store};
\node[choice, below=1.2mm of checks.south west, anchor=north west, align=left,
      text width=81mm] (disp)
  {\emph{above 64 capabilities:} a dispatcher replaces per-tool schemas,
   holding the re-sent schema at $O(1)$};
\end{tikzpicture}
\vspace{-3.5mm}
\caption{The Capability Builder through the 1000 Genomes run, where the tracer install becomes a tool because its link flag never changes and the annotation choice stays a skill because it does.}
\vspace{-4mm}
\label{fig:c1}
\end{figure}
This component turns a solution produced during a task into a reusable capability. Without it, an agent produces only an answer and writes nothing executable; with it, the knowledge gained during the task can be reused by subsequent tasks.
Three challenges follow.
First, because a module that fails to import cannot be used by a later task, which has none of the context that produced it, a capability must be \emph{correct at the moment it is written}.
Second, because an agent that cannot tell whether a tool already exists may write another one, the store must avoid \emph{near-duplicates}: two spellings of the same procedure can make capabilities harder to find and reuse.
Third, because judgment that cannot be frozen into code is what a skill must carry, while deterministic steps described in prose may be performed differently next time, the \emph{split between code and prose} must be explicit.\looseness=-1

Figure~\ref{fig:c1} follows the component through the 1000 Genomes run of Section~\ref{sec:motivation}, showing what becomes a tool and what remains a skill. The tracer install the baseline fixed by swapping its whole toolchain becomes an \ac{MCP} tool that injects the missing link flag, because that flag is the same on this system every time, while the choice of which functions carry the I/O worth annotating remains a skill because it requires judgment.
Each capability is a single Python module exposing \texttt{register(mcp)}, filed in a layer's \texttt{capabilities/} directory, so it travels with that layer and needs no installation step for the next person who clones the project.
Writing one goes through \texttt{capability\_new}, which validates at write time rather than load time, compiling the source, requiring the \texttt{register} entry point, rewriting machine-specific paths out, and refusing a duplicate by fingerprinting the module with docstrings, comments, and the function's name removed, so a twin under another name is recognised across every layer.
The judgment the tool cannot encode becomes skill blocks, \texttt{SKILL.md} carrying instructions, rules, and references and \texttt{LESSONS.md} carrying pitfalls and lessons, each addressed by a stable label that is never recycled, so a rule can be cited from another skill the way a foreign key cites a row.\looseness=-1

A skill file grows with every task that touches it, accumulating history, corner cases, and lessons a later task has no interest in, while a block is the pointed answer to one question, so the two or three blocks a task needs cost a fraction of its files.
Retrieval therefore reads the store as a query reads a database, consulting an index to open only the rows that answer it rather than scanning every table, and it returns blocks rather than files, scoring a catalogue of subjects and coverage rows, folding matched rows back into the query, and returning at most five skills of at most four blocks each, with a relevance floor that lets it return nothing rather than something loosely related.
A language-aware graph backs that catalogue, indexing capabilities, skill blocks, and research records as nodes so a query names the few objects worth opening, which matters because locating costs a few hundred tokens while reading the files it names costs tens of thousands.
Retrieval costs a linear scan of that index, $O(S)$ in skills for the catalogue pass and $O(B)$ in the blocks of the surviving skills, with no embedding index to maintain and no state that can fall out of step with the files.
The dominant cost is instead the tool schemas a harness re-sends on every request, which reached about 95K tokens at roughly 1{,}150 learned tools, so above 64 capabilities \phronix{} publishes a dispatcher pair rather than individual tools and returns the per-capability schema cost to $O(1)$, at the price of one level of indirection.\looseness=-1

\subsection{Tracker: Automatic Tracking and Knowledge Creation}
\label{sec:design-c2}

\begin{figure}[t]
\centering
\begin{tikzpicture}[phx]
\node[stage, minimum width=19.5mm] (obs) {\texttt{observe}};
\node[stage, right=1.5mm of obs, minimum width=19.5mm] (group) {group};
\node[stage, right=1.5mm of group, minimum width=19.5mm] (distil) {distil};
\node[stage, right=1.5mm of distil, minimum width=19.5mm] (promote) {promote};

\node[bad, below=3mm of obs.south, anchor=north, align=left, text width=18mm] (o1)
  {\emph{worked}=false build fails on the default compiler};
\node[good, below=1.2mm of o1.south, anchor=north, align=left, text width=18mm] (o2)
  {\emph{worked}=true it links with \texttt{-lstdc++fs}};
\node[art, below=3mm of group.south, anchor=north, align=left, text width=18mm] (g1)
  {both land under \emph{tooling}, not under this workflow};
\node[art, below=3mm of distil.south, anchor=north, align=left, text width=18mm] (d1)
  {one lesson: this system needs the flag};
\node[graph, below=3mm of promote.south, anchor=north, text width=18mm] (p1)
  {written to the project layer, and the \textbf{graph} is marked stale};

\draw[flow] (obs) -- (group);
\draw[flow] (group) -- (distil);
\draw[flow] (distil) -- (promote);
\draw[flow] (obs.south) -- (o1.north);
\draw[flow] (group.south) -- (g1.north);
\draw[flow] (distil.south) -- (d1.north);
\draw[flow] (promote.south) -- (p1.north);

\node[choice, below=2mm of o2.south west, anchor=north west, align=left,
      text width=83mm] (note1)
  {\emph{nobody asked for either entry:} reporting is a non-blocking call the
   harness makes on every interaction};
\node[choice, below=1.2mm of note1.south west, anchor=north west, align=left,
      text width=83mm] (note2)
  {\emph{filed by kind, not by task:} the next project that builds this tracer
   retrieves the lesson without knowing genomics produced it};
\end{tikzpicture}
\vspace{-3.5mm}
\caption{The Tracker through the same run, where the failed build and the one that followed are reported automatically, grouped by kind of operation rather than by workflow.}
\vspace{-4mm}
\label{fig:c2}
\end{figure}
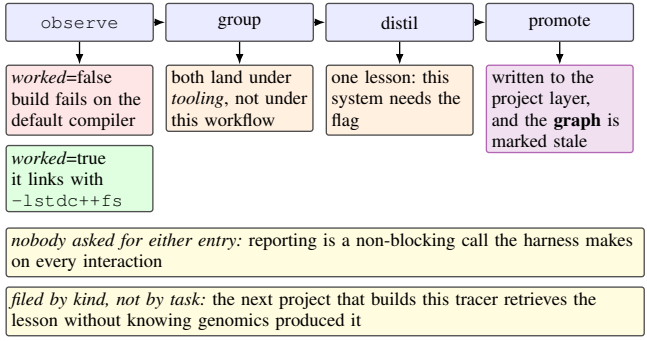

This component removes the user from the learning loop, because the evidence that a step worked or failed exists only while the task runs, and a harness that misses it can only ask the user afterwards what was worth keeping.
Three challenges follow.
First, since a record that costs a model call is made only where someone already suspects it matters, which is where it is least needed, reporting must be \emph{cheap enough to be unconditional}.
Second, because a lesson filed under the task that produced it helps that task alone, what is captured must be \emph{filed so that it generalises}.
Third, as a task that pauses to write skills pays for learning in the latency of the work itself, distillation must not sit on the \emph{critical path}.\looseness=-1

Figure~\ref{fig:c2} follows the component through the same run, where the tracer build that failed under the site's default compiler and the build that succeeded once the link flag was added are both recorded as they happen, without anyone recognising in advance that this pair was the valuable one, and filing them under the kind of operation rather than this workflow is what lets the next project that builds the tracer retrieve the lesson.
Every interaction is reported through \texttt{observe}, which records what was attempted, whether it worked, a short detail, the stage, and the kind of operation, appending it to a log while publishing it on a socket that never blocks and never fails the caller, so a run without a listener degrades to the file alone.
Observations are filed by \emph{kind of operation} rather than by task, over a fixed vocabulary such as subprocess execution, tests, patching, and configuration, so what is learned about running subprocesses accumulates in one place across every task that ran one.
Grouping is arithmetic rather than model-driven, normalising each report into a shape with digits and paths removed and suffixes folded, then merging shapes whose token overlap passes a fixed threshold, which lets the cycle run every ten seconds without an agent invocation.\looseness=-1

Learning must never delay the task that produces it.
Distillation therefore runs off the critical path in a background learner on a one-minute cycle with a bounded budget, partitioned so one worker owns one kind of operation and two can never write contradictory lessons into one file.
Every write marks the knowledge graph stale rather than rebuilding it, and a background clock refreshes the index, keeping the rebuild off the query path where it would cost five times the latency of a warm lookup.
The per-cycle cost is linear in the observations since the last cycle and quadratic only within the bounded candidate set it considers, so tracking adds a constant overhead per tool call and no term that grows with the store.\looseness=-1

Promotion into the durable layer makes a task's findings outlive it, and happens after every task rather than at the end of a run.
A task's end is the first point at which the outcome is known, where a solution has either worked or failed in a way that names what was missing, which is the evidence promotion should gate on.
This matters especially for a killed run, since the tasks it finished carry graded outcomes that run-level promotion would discard.
Promotion merges rather than mirrors, adding and updating files but never deleting, each write published atomically so concurrent tasks cannot tear each other's output, at a cost linear in the files a task changed.\looseness=-1

\subsection{Researcher: Continual-Learning-Based Research}
\label{sec:design-c3}

\begin{figure}[t]
\centering
\begin{tikzpicture}[phx]
\node[choice, minimum width=19.5mm] (gate) {gate};
\node[stage, right=1.5mm of gate, minimum width=19.5mm] (plan) {plan};
\node[stage, right=1.5mm of plan, minimum width=19.5mm] (try1) {attempt 1};
\node[stage, right=1.5mm of try1, minimum width=19.5mm] (try2) {attempt 2};

\node[graph, below=3mm of gate.south, anchor=north, text width=18mm] (g1)
  {\textbf{graph query}: nothing stored on annotating a data loader, so it is a gap};
\node[art, below=3mm of plan.south, anchor=north, align=left, text width=18mm] (p1)
  {edits are deterministic; which paths carry I/O is judgment};
\node[bad, below=3mm of try1.south, anchor=north, align=left, text width=18mm] (t1)
  {fails, but reaches a later step than before};
\node[good, below=3mm of try2.south, anchor=north, align=left, text width=18mm] (t2)
  {earned by that progress, and it solves the task};

\draw[flow] (gate) -- (plan);
\draw[flow] (plan) -- (try1);
\draw[flow] (try1) -- node[lbl,pos=0.5,yshift=2.6mm]{+1} (try2);
\draw[flow] (gate.south) -- (g1.north);
\draw[flow] (plan.south) -- (p1.north);
\draw[flow] (try1.south) -- (t1.north);
\draw[flow] (try2.south) -- (t2.north);

\node[choice, below=2mm of g1.south west, anchor=north west, align=left,
      text width=83mm] (note1)
  {\emph{progress is a position, not a score:} a later step reached earns a
   try, up to two, and a changed failure buys one more};
\node[choice, below=1.2mm of note1.south west, anchor=north west, align=left,
      text width=83mm] (note2)
  {\emph{stagnation ends it:} the same failure twice, or the same patch again,
   stops the task rather than spending the budget};
\end{tikzpicture}
\vspace{-3.5mm}
\caption{The Researcher through the annotation step, where the lookup finds a gap, the plan separates deterministic edits from judgment, and a failing attempt still reaches a later step.}
\vspace{-4mm}
\label{fig:c3}
\end{figure}
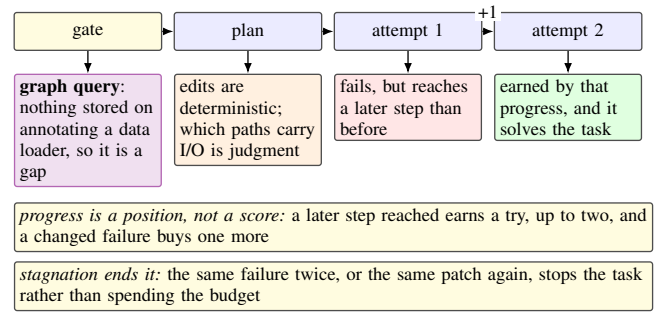
This component decides what happens when the store does not contain the answer, the case that determines whether an agent grows or merely applies what it has, since a harness that simply retries spends its budget resampling a model that has already failed.
Three challenges follow.
First, because a model is wrong about what it knows in exactly the cases where that judgment matters, the decision to research must not rest on the \emph{model's own confidence}.
Second, since a task that has not yet passed its tests looks identical on every attempt if success is the only signal, progress must be \emph{measurable while a task is still failing}.
Third, because an agent that keeps trying while nothing changes converts a failure into an expensive failure, effort must be \emph{withdrawn as decisively as it is granted}.\looseness=-1

Figure~\ref{fig:c3} follows the component through the annotation step of the same run, where the lookup finds nothing stored about annotating a Python data loader and so treats the task as a gap to research rather than one to retry, and where the attempt that fails still reaches a later step of the plan, which earns the attempt that finishes the task.
Research runs once before the first attempt, gated by a deterministic coverage check that counts the catalogue blocks matching the task and treats fewer than two as a gap, so a task the store covers costs a lookup rather than a research pass.
That check and the search for prior research are graph queries, so a task inherits what an earlier one established about the same subject without the model reading either record.
A plan states the steps that solve the class of problem, every step tagged deterministic or judgment-based and untagged steps rejected, so the plan itself decides what becomes an \ac{MCP} tool and what becomes a skill block.
A figure of merit is fixed before any approach is measured, approaches carry their source, and a status such as working or falsified is reachable only by measurement on held-out instances rather than by argument.\looseness=-1

Progress within a failing task is measured over a range rather than as a binary pass or fail, because a plan is an ordered method and the step an attempt reaches says how far the agent got toward its stated goal.
Failing at step four after failing at step two is therefore a task getting closer, while the count of passing tests stays at zero and the binary signal reports nothing.
The budget follows that signal, starting at three attempts, granting up to two more when an attempt reaches a step no earlier attempt reached, and one more when the failure signature changes, the evidence that the idea moved rather than the patch.
It is withdrawn on the same evidence, since two consecutive identical failure signatures, or a repeated patch, count as a stall and end the task, while attempts that never ran are void.
The loop therefore bounds a task at five attempts and terminates in two once nothing changes, adding a constant number of agent invocations per task rather than a term that grows with the store or the task's length.\looseness=-1

\subsection{Implementation}
\label{sec:design-impl}

\phronix{} is implemented in Python as an \ac{MCP} server publishing 84 tools across 16 services, with retrieval registered first because tool order influences what a model reaches for.
The store resolves over three layers, bundled inside the package, user under the home directory, and project inside the repository, where a higher layer shadows a lower one wholesale rather than merging block by block, so no document exists that no author wrote.
Writes default to the project layer, placing learning in the same commit and review as the code it describes, and a one-minute watcher registers capabilities appearing while the server runs, so a tool the learner writes becomes callable without a restart.\looseness=-1

That one store serves every run mode and every harness.
Cold- and warm-start runs differ only in what the store holds at the start, a cold store built from the bundled layer alone and a warm one laid over it from the promoted layer, and the framework records which mode took effect because a cold run reporting carried capabilities is not the experiment it claims to be.
One server and one store serve every harness, rendering agent definitions, permissions, and prompt hooks from one neutral definition in which agents declare a capability level rather than a model name, so the same definition runs on Claude Code, OpenCode, Codex, and Copilot and a model upgrade is a one-line change.
Learning is shared at three mechanisms, the project layer committed with the repository, a promotion root collecting what a run produced so research is not paid for twice, and a contribute step promoting skills, agents, capabilities, and verifiers upstream to everyone who installs \phronix{}, which refuses anything its privacy scan flags and excludes memory as the most project-specific tier.\looseness=-1

\section{Evaluation}
\label{sec:evaluation}

We evaluate \phronix{} in four stages, first describing our methodology and establishing the overall solve rate every later result is read against.
We then evaluate each of the three concepts, benchmark the end-to-end token cost of solving a task, and apply \phronix{} to an HPC I/O performance-characterization pipeline.\looseness=-1

\subsection{Methodology}
\label{sec:eval-methodology}

\subsubsection{Testbed}
We run all experiments on \cluster{}, whose nodes each have a 48-core AMD Rome processor, 256~GB of memory, and a Lustre parallel file system.
Each SWE-bench run occupies 8 nodes and executes every attempt in its own rootless Podman container built from the task's official SWE-bench image, so no attempt can observe another.\looseness=-1

\subsubsection{Harness and Models}
Both arms of every comparison run on the Claude Code~\cite{anthropic_claude_code} harness (v2.1.246) at low reasoning effort, so extended reasoning does not become a second variable alongside \phronix{}.
We drive three models of increasing capability and cost, gpt-5.4, gpt-5.5, and gpt-5.6-sol, reached through an OpenAI-compatible institutional gateway.
The use case in Section~\ref{sec:eval-usecase} runs the same harness on gpt-5.5, chosen because it solved a task at the lowest cost of the three once its store was populated, at \$1.05--\$1.23 per solved task against \$1.26 for gpt-5.4 and \$1.35 for gpt-5.6-sol.\looseness=-1

\subsubsection{Workloads}
Software-engineering tasks suit this evaluation, because each is a real repository change whose ground-truth tests decide objectively whether the agent solved it, and solving one exercises the whole loop of reading a codebase, changing it, and validating the change.\looseness=-1
We evaluate on 30 tasks from SWE-bench Lite~\cite{jimenez_swe-bench_2024}, selected deterministically so every repository is represented and the tasks span patches of 1 to 76 lines with regression surfaces of 0 to 1689 tests.
Every model solves the same 30 in the same order, letting a later task benefit from the capabilities earlier ones created.
As 12 repositories are too many to compare individually, we group them by domain into the five categories of Table~\ref{tab:solve-rate}.
The use case in Section~\ref{sec:eval-usecase} adds the 1000 Genomes Project workflow~\cite{1000genomes_global_2015}, executed with Pegasus~\cite{deelman_pegasus_2015} and instrumented with DFTracer~\cite{devarajan_dftracer_2024} for I/O tracing.\looseness=-1

\begin{table*}[t]
\centering
\caption{Tasks solved over tasks attempted, pooled across runs, for the 30 SWE-bench Lite tasks in five categories.}
\vspace{-2mm}
\label{tab:solve-rate}
\small
\begin{tabular}{llcccccccccc}
\toprule
\multicolumn{3}{c}{} & \multicolumn{3}{c}{gpt-5.4} & \multicolumn{3}{c}{gpt-5.5} & \multicolumn{3}{c}{gpt-5.6-sol} \\
\cmidrule(lr){4-6}\cmidrule(lr){7-9}\cmidrule(lr){10-12}
Category & Repositories & Tasks & Base & Cold & Warm & Base & Cold & Warm & Base & Cold & Warm \\
\midrule
Web \& HTTP & django, flask, requests & 9 & 13/45 & 13/18 & 20/22 & 18/63 & 21/27 & 32/36 & 51/72 & 24/27 & 40/45 \\
Dev tooling & pytest, pylint, sphinx & 6 & 7/30 & 9/12 & 15/15 & 11/42 & 12/18 & 23/24 & 36/48 & 18/18 & 30/30 \\
Scientific & astropy, sympy & 5 & 10/25 & 4/10 & 10/12 & 10/35 & 10/15 & 15/20 & 17/40 & 11/15 & 19/25 \\
Data \& ML & xarray, scikit-learn & 5 & 3/25 & 6/10 & 11/14 & 7/35 & 9/15 & 14/20 & 18/40 & 11/15 & 20/25 \\
Visualization & matplotlib, seaborn & 5 & 10/25 & 8/10 & 14/14 & 11/35 & 15/15 & 20/20 & 33/40 & 15/15 & 25/25 \\
\midrule
\multicolumn{2}{l}{\textbf{Solved (\%)}} & 30 & \textbf{29} & \textbf{67} & \textbf{91} & \textbf{27} & \textbf{74} & \textbf{87} & \textbf{65} & \textbf{88} & \textbf{89} \\
\bottomrule
\end{tabular}

\vspace{-4mm}
\end{table*}

\subsubsection{Run Protocol}
We compare three arms sharing the same tasks, containers, and model.
The \emph{baseline} runs the harness without \phronix{}, a \emph{cold-start} run uses \phronix{} from an empty capability store, and a \emph{warm-start} run runs it from the store the cold-start runs promoted.
Every run pairs the baseline with one \phronix{} arm, so they differ only in whether \phronix{} is deployed.
We charge the baseline a fixed five attempts per task, estimated from its single measured attempt, because its attempts carry nothing between them by design, which is precisely what \phronix{} adds.
Both \phronix{} arms instead vary attempts with progress, starting at three, earning up to two more when an attempt progresses further than any before it, and stopping after two that fail identically.
Each attempt has a 30-minute timeout, and we repeat every configuration up to five times, discarding runs in which the wall-time limit rather than the model ended at least half the tasks and dropping both arms to keep the comparison paired.\looseness=-1

\subsubsection{Metrics}
\label{sec:metrics}
A task is \emph{solved} when every FAIL\_TO\_PASS test of the official test patch passes and no PASS\_TO\_PASS test regresses, and the \emph{solve rate} is the fraction of attempted tasks solved, pooled over runs.
The \emph{first-try rate} is the fraction of tasks a cold-start run solved that a warm-start run then solved on its first attempt.
We further count the skills and \ac{MCP} tools created and reused, the entries the tracker records, and the attempts and tokens a task consumes, including prompt-cache reads.
Every value is a mean over repeated runs, with error bars showing the sample standard deviation.\looseness=-1

\subsection{Overall Solve Rate}
\label{sec:eval-solve}


Table~\ref{tab:solve-rate} reports how many tasks each arm solves for each model.
The baseline solves 29\%, 27\%, and 65\%, which measures what each model achieves on its own.
A cold-start run raises these to 67\%, 74\%, and 88\%, because capabilities that earlier tasks create in the same run already help later ones.
A warm-start run raises them to 91\%, 87\%, and 89\%, so all three converge on roughly the same rate once they start from a populated store.
The gain holds in every category, where \phronix{} solves 96--100\% of the Dev tooling and Visualization tasks and 70--83\% of the harder Scientific and Data \& ML tasks, and is largest for gpt-5.4 on Data \& ML, rising from 12\% to 79\%.
\phronix{} therefore lets the two cheaper models match the solve rate of the most capable and costly one.\looseness=-1

\subsection{Internal Evaluations}
\label{sec:eval-internal}

\subsubsection{Deterministic but Intelligent Capabilities}
\label{sec:eval-capabilities}

\begin{figure}[b]
    \vspace{-4mm}
\centering
\includegraphics[width=\linewidth]{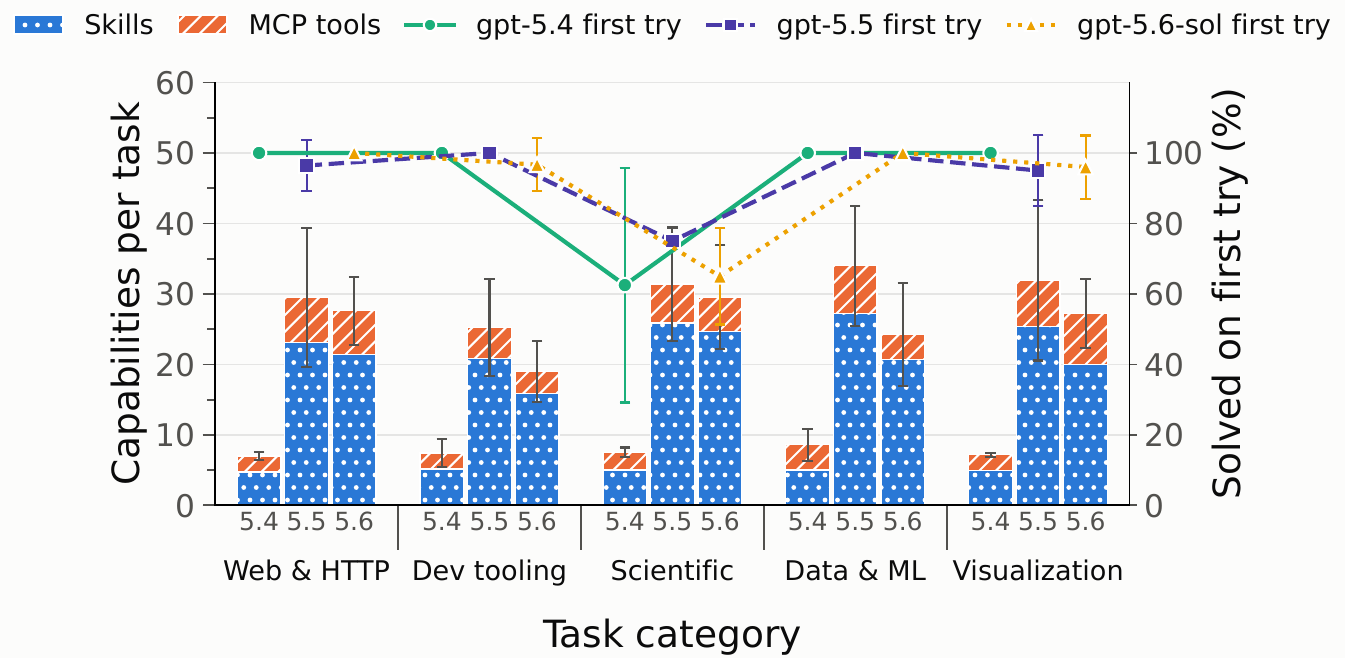}
\vspace{-3.5mm}
\caption{Every model pairs \ac{MCP} tools with skills in every category, and a warm-start run solves 95\% of previously solved tasks in one attempt.}

\label{fig:e1}
\end{figure}

To show that capabilities are deterministic and intelligent, we confirm that \phronix{} builds an \ac{MCP} tool together with skills for every kind of task, and that the result solves a recurring task without rediscovering its solution.
Our test runs the 30 tasks cold-start and then warm-start, so the second run begins from the store the first promoted.
We measure the skills and \ac{MCP} tools a cold-start run builds, and the fraction of tasks a warm-start run solves on its first attempt among those previously solved by the cold-start run.
Figure~\ref{fig:e1} plots task category on the x axis with one bar per model, capabilities built per task on the left y axis split into skills and \ac{MCP} tools, and that first-attempt rate on the right y axis as one line per model.\looseness=-1

Every model builds both kinds in every category, gpt-5.4 building 7--9 capabilities per task against 19--34 for the other two, skills outnumbering \ac{MCP} tools two to four times because the tool encodes what can be coded and the skills carry the judgment around it.
The warm-start lines confirm that these capabilities solve the task, at 94.8\%, 93.5\%, and 93.3\% of previously solved tasks on the first attempt.
Four of five categories reach 95--100\% with a standard deviation of at most 10 points, and Data \& ML reaches 100\% in every run of every model, which shows the solution is replayed rather than searched for again.
Scientific is the exception at 62--75\% with the widest spread, because no model reproduces the two astropy tasks on its first attempt.
gpt-5.4 is as deterministic as the other two while building three to four times fewer capabilities per task, so determinism follows from the tool-and-skill pairing rather than the number of capabilities or the model.

\noindent\textbf{Key finding:} \emph{Capabilities that pair an \ac{MCP} tool with skills solve a recurring task on the first attempt in 93--95\% of cases across every model, which removes solution non-determinism.}\looseness=-1

\subsubsection{Automatic Tracking and Knowledge Creation}
\label{sec:eval-tracking}

\begin{figure}[b]
    \vspace{-4mm}
\centering
\includegraphics[width=\linewidth]{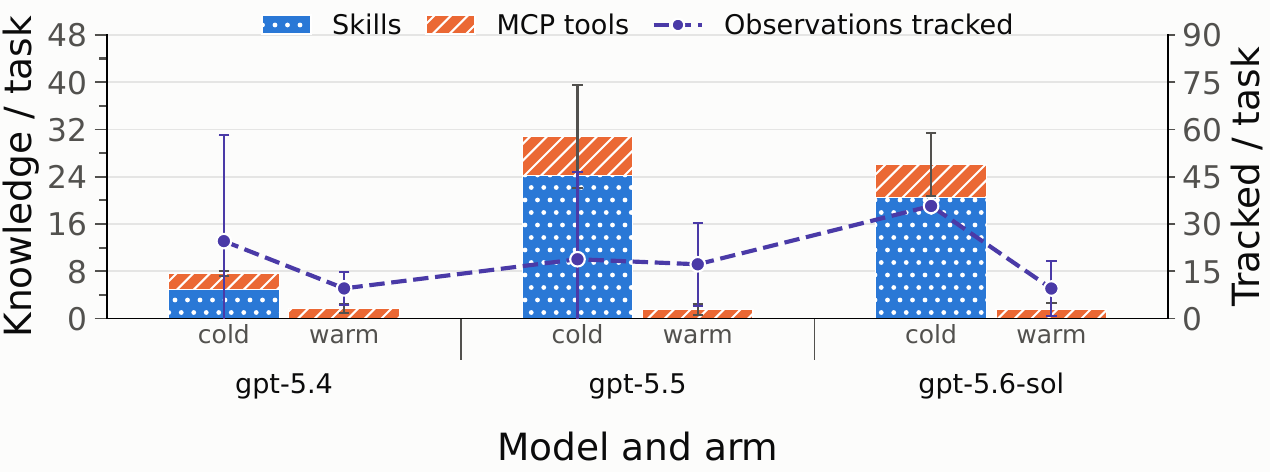}
\vspace{-3.5mm}
\caption{A cold-start run distills 7.6--30.8 capabilities per task unasked (bars, left), while a warm-start run keeps tracking (line, right) but creates under two.}

\label{fig:e2}
\end{figure}

To show that \phronix{} learns without being instructed, we confirm that it records what happens while a task is solved and turns those records into durable knowledge on its own.
Our test runs the same 30 tasks with no instruction to remember anything, so every observation and capability created during these runs is one \phronix{} decided to create.
We measure the observations recorded and the skills and \ac{MCP} tools distilled from them in both arms.
Figure~\ref{fig:e2} plots model and arm on the x axis, knowledge per task on the left y axis split into skills and tools, and observations tracked per task on the right.\looseness=-1

A cold-start run creates 7.6, 30.8, and 26.1 capabilities per task, or 229, 924, and 782 per run, filed under 16 to 58 distinct topics.
A warm-start run creates fewer than two per task and almost no skills, while its tracker keeps recording, which shows knowledge creation diminishes when the capability a task needs already exists rather than when tracking stops.
The tracked volume varies widely between runs of one configuration because observations are attributed by timestamp, so a run whose store carried earlier timestamps reports far fewer, and we read the line as evidence that tracking continues rather than as a precise volume.
The knowledge counts do not depend on that attribution, and they show the conversion from observation to capability is what differs between the arms.
Because no user asked for any of this, the tracked record, not the user's memory of what worked, is what \phronix{} learns from.

\noindent\textbf{Key finding:} \emph{Validating every distilled observation with a tool-only check rather than with the model converts an agent's own execution into 7.6--30.8 durable capabilities per task with no user instruction, and creation falls below two per task once the needed capability exists.}\looseness=-1

\subsubsection{Continual-Learning-Based Research}
\label{sec:eval-research}

\begin{figure}[t]
\centering
\includegraphics[width=\linewidth]{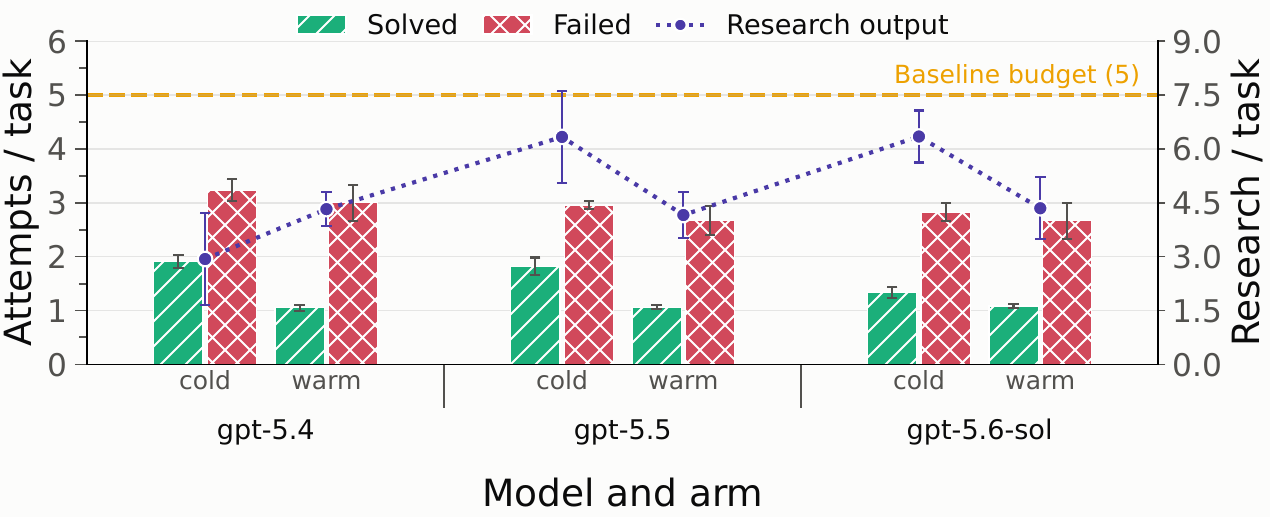}
\vspace{-3.5mm}
\caption{Research prices attempts by progress, a solved task costing 1.05--1.91 attempts against the baseline's five and a stagnating one stopping near three. Whiskers show the standard deviation.}
\vspace{-4mm}
\label{fig:e3}
\end{figure}

To show that research makes a failing task productive rather than expensive, we confirm that \phronix{} spends attempts where they make progress, stops where they do not, and leaves knowledge behind either way.
Our test compares the attempts a task consumes in both arms against the fixed five-attempt budget charged to the baseline, which spends that budget whatever happens while \phronix{} lets progress decide.
We measure attempts separately for solved and failed tasks, because averaging the two would hide both, together with the topics researched and the \ac{MCP} tools research produced.
Figure~\ref{fig:e3} plots model and arm on the x axis, attempts per task by outcome on the left y axis with the baseline budget as a dashed rule, and research output per task on the right y axis.\looseness=-1

A solved task costs 1.91, 1.82, and 1.33 attempts cold-start and 1.05, 1.06, and 1.08 warm-start, a fifth to two-fifths of the budget the baseline spends on the same task, so research finds the solution early rather than resampling the model.
A failed task stops at 2.67 to 3.23 attempts rather than exhausting the budget, because two attempts that fail identically end it, so stagnation costs less than the baseline's fixed five.
Research output continues in both arms, 2.93--6.34 per task cold-start and 4.16--4.35 warm-start, so even an unsolved task leaves topics and tools a later task can use.
The pattern holds across all three models, so pricing attempts by progress is a property of the research loop rather than of a particular model.
A cold-start run spends more attempts than a warm-start one on the tasks both solve, which is the cost of discovering a capability that a warm-start run then replays for roughly one attempt.

\noindent\textbf{Key finding:} \emph{Per-attempt improvement is a workable convergence criterion for the predictor--validator--corrector loop, since research spends 1.05--1.91 attempts on a solved task against a five-attempt budget and stops a stagnating one near three, while still producing 2.9--6.3 research outputs per task.}\looseness=-1

\subsection{End-to-End Benchmarking}
\label{sec:eval-tokens}

\begin{figure}[t]
\centering
\includegraphics[width=\linewidth]{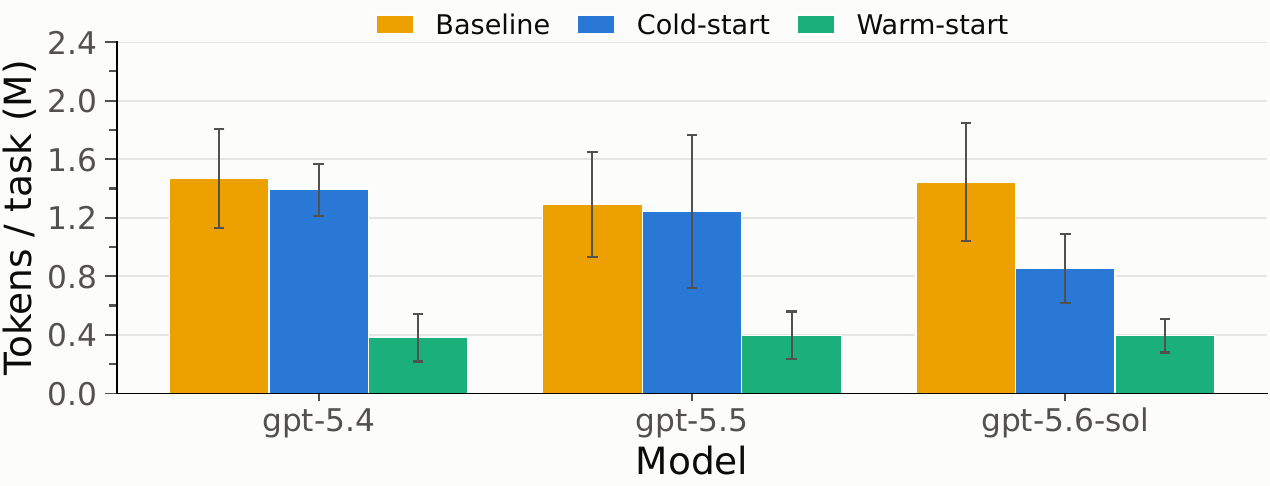}
\vspace{-3.5mm}
\caption{A warm-start run consumes 0.38--0.40M tokens per task, three to four times less than one baseline attempt, and a cold-start run costs no more than the baseline while solving far more tasks. Whiskers show the standard deviation.}
\vspace{-4mm}
\label{fig:e4}
\end{figure}

To show that learning pays for itself end to end, we confirm that the capabilities \phronix{} builds cost less to apply than solving the same task without them, since a harness carrying a large context could otherwise trade tokens for its higher solve rate.
Our test prices a whole task rather than a single attempt, comparing the baseline against both arms on the same 30 tasks and models.
We measure every token a task consumes, including the prompt-cache reads that are 88--92\% of them, reporting cost per task so arms attempting a task different numbers of times stay comparable.
Figure~\ref{fig:e4} plots model on the x axis and tokens per task in millions on the y axis, one bar per arm.\looseness=-1

A warm-start run consumes 0.38M, 0.40M, and 0.39M tokens per task, 3.9, 3.2, and 3.7 times less than the 1.47M, 1.29M, and 1.44M a single baseline attempt consumes, and 16 to 19 times less than the five-attempt budget the baseline is charged.
That cost is nearly identical across the three models, so the price of applying a capability is set by the capability rather than the model running it.
A cold-start run costs 1.39M, 1.24M, and 0.85M tokens per task, at or below a single baseline attempt, so building capabilities while solving a task does not make it more expensive even before any reuse.
Because 88--92\% of these tokens are prompt-cache reads, billed well below fresh input, a comparison in currency would favour \phronix{} further.
Taken with Table~\ref{tab:solve-rate}, \phronix{} solves three times as many tasks as the baseline for gpt-5.4 on a quarter of the tokens, so the store lowers cost and raises the solve rate together.

\noindent\textbf{Key finding:} \emph{A warm-start run solves a task for 0.38--0.40M tokens, three to four times less than a single baseline attempt and 16 to 19 times less than its charged retry budget, while a cold-start run costs no more than the baseline.}\looseness=-1

\subsection{HPC Use Case: The 1000 Genomes Workflow}
\label{sec:eval-usecase}

\begin{figure}[t]
\centering
\includegraphics[width=\linewidth]{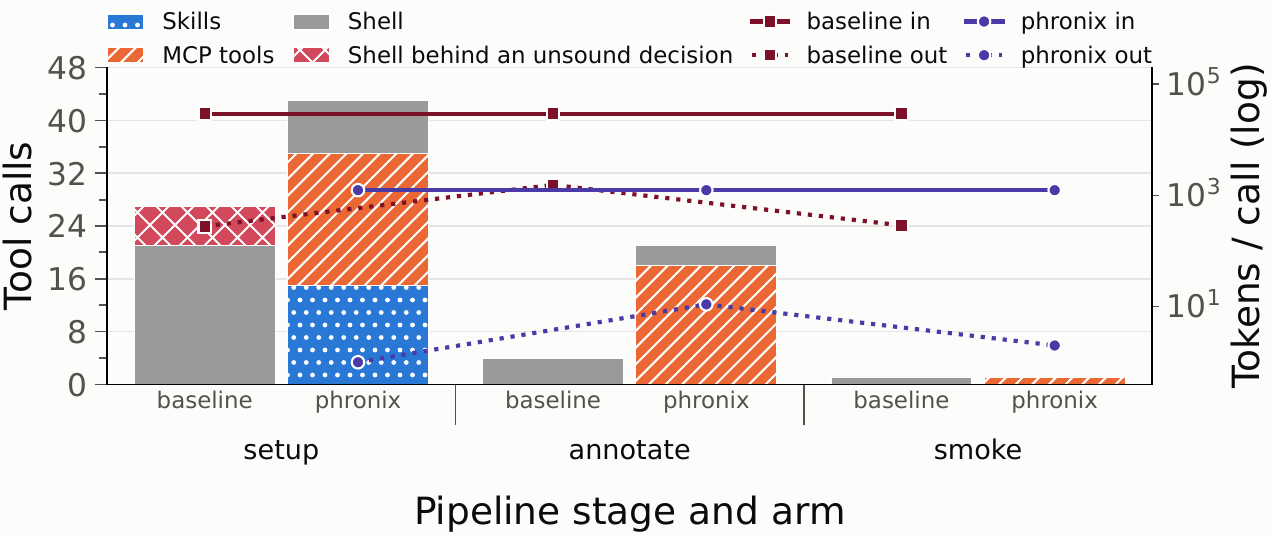}
\vspace{-3.5mm}
\caption{\phronix{} instruments the 1000 Genomes workflow through capabilities it built, while every baseline call is shell the model composed. Red marks the six behind unsound decisions.}
\vspace{-4mm}
\label{fig:e5}
\end{figure}


To show that capability-centric execution transfers beyond software-engineering benchmarks, we confirm that driving a real HPC task through \ac{MCP} tools and skills, rather than shell composed on the spot, makes its outcome deterministic and correct.
Our test instruments the 1000 Genomes workflow~\cite{1000genomes_global_2015} with DFTracer~\cite{devarajan_dftracer_2024}, which requires installing the tracer, annotating the read and per-individual write paths, and running a smoke test that exercises the annotated application and emits a trace, on the Claude Code harness driving gpt-5.5 in both arms.
We measure how many calls go through an \ac{MCP} tool or skill rather than shell, excluding those declined at a permission prompt, what each costs in input and output tokens, and how sound each decision behind the finished workflow was, scoring that workflow against 11 correctness checks backed by evidence from the run rather than the agent's own report.
Figure~\ref{fig:e5} plots pipeline stage and arm on the x axis, tool calls on the left y axis split into skills, \ac{MCP} tools, shell, and failed shell, and input and output tokens per call on a logarithmic right y axis, while Table~\ref{tab:e5-decisions} scores every decision.\looseness=-1

\phronix{} drives the pipeline through the capabilities it built, with 41 \ac{MCP} tool calls and 15 skill invocations, and the shell it issues carries out steps a skill prescribes where the judgment cannot be frozen into static code.
Every baseline call instead invokes a shell command the model wrote for itself while working out the problem, so the same pipeline runs on whatever it composed at that moment.
That difference decides the outcome, as \phronix{} passes all 11 correctness checks while the baseline passes 6, reaching two of the remaining five only through a workaround and failing three outright.
Both arms complete the smoke test, so the gap is not whether the annotated application runs but whether the instrumentation is correct, and the three failures are what an agent cannot see in its own output, since the baseline never lints or validates its annotations and leaves the outer compression function without a span.
Passing a check is not the same as deciding well, and Table~\ref{tab:e5-decisions} shows \phronix{} sound on all 11 decisions and the baseline on 5, from downgrading the interpreter after a dependency it never needed failed to build, to restructuring the application so its write loop could be annotated at all.
The workarounds tell the same story, as the baseline diagnosed a missing \texttt{-lstdc++fs} link flag exactly and then swapped the whole toolchain instead of applying it, whereas \phronix{} injects that flag from an \ac{MCP} tool built in an earlier run and installs the tracer without incident.
The token lines price this per action, the baseline spending 29{,}605 input and 279--1{,}549 output tokens per call against 1{,}246 and 1--11 for \phronix{}, because a named tool takes a short structured request whereas a baseline call must write out the shell to run.
Driving the pipeline through tools therefore leaves less to generate and less to get wrong, and six baseline setup calls, drawn in red in Figure~\ref{fig:e5}, rebuilt the environment and swapped the toolchain.

\begin{table}[t]
\centering
\caption{Decisions behind the finished workflow, \textcolor{okgreen}{green} where sound and \textcolor{badred}{red} where not. Only 5 of the baseline's 11 are sound.}
\vspace{-2mm}
\label{tab:e5-decisions}
\footnotesize
\setlength{\tabcolsep}{3pt}\begin{tabular}{@{}p{0.10\linewidth}p{0.21\linewidth}p{0.29\linewidth}p{0.29\linewidth}@{}}
\toprule
Stage & Decision & Baseline & \phronix{} \\
\midrule
setup & Python interpreter & \textcolor{badred}{downgraded to 3.9.12} & \textcolor{okgreen}{kept the session interpreter} \\
\cmidrule(l){2-4}
 & Tracer build toolchain & \textcolor{badred}{swapped the toolchain} & \textcolor{okgreen}{added the missing link flag} \\
\cmidrule(l){2-4}
 & Tracer release version & \textcolor{okgreen}{2.1.1.post36.dev0} & \textcolor{okgreen}{2.1.1.post36.dev0} \\
\cmidrule(l){1-4}
annotate & Annotation placement & \textcolor{badred}{restructured the app} & \textcolor{okgreen}{annotated in place} \\
\cmidrule(l){2-4}
 & Annotation checking & \textcolor{badred}{none} & \textcolor{okgreen}{linted and validated} \\
\cmidrule(l){2-4}
 & Write-path coverage & \textcolor{okgreen}{2504 write events} & \textcolor{okgreen}{2504 write events} \\
\cmidrule(l){1-4}
smoke & Smoke test & \textcolor{okgreen}{ran to completion} & \textcolor{okgreen}{ran to completion} \\
\cmidrule(l){1-4}
trace & Tar-step span coverage & \textcolor{badred}{outer function unspanned} & \textcolor{okgreen}{spanned both levels} \\
\cmidrule(l){2-4}
 & Trace produced & \textcolor{okgreen}{2512 events} & \textcolor{okgreen}{2520 events} \\
\cmidrule(l){1-4}
other & Waiting for long steps & \textcolor{badred}{polled the filesystem} & \textcolor{okgreen}{waited on the tool} \\
\cmidrule(l){2-4}
 & Input staging & \textcolor{okgreen}{copied and verified} & \textcolor{okgreen}{copied} \\
\midrule
\multicolumn{2}{@{}l}{\textbf{Decisions sound}} & \textbf{5/11} & \textbf{11/11} \\
\bottomrule
\end{tabular}

\vspace{-4mm}
\end{table}

\noindent\textbf{Key finding:} \emph{Predicting with capabilities, validating with tool-only checks, and correcting what they find, rather than letting the model compose and judge its own shell, makes the outcome deterministic, instrumenting the application correctly on all 11 checks against 6 for the baseline while generating 25--140$\times$ fewer output tokens per call.}\looseness=-1

\section{Discussion}
\label{sec:discussion}

Our results raise two questions the measurements do not settle, and both mark the boundary between what a capability encodes and what the environment around it supplies.\looseness=-1

\noindent\textbf{Transfer across models and harnesses.}
A cold-start run pays the full price of discovering a capability while a warm-start run applies it for a fifth of the attempts and a third of the tokens, which argues for populating the store once with an expensive model and drawing on it with a cheap one, as gpt-5.4 and gpt-5.5 do in reaching 91\% and 87\% from a store their own family produced.
Across families the transfer is less clean in our experience, because a skill encodes not only what to do but how the model it was written for prefers to be told.
The research loop of Section~\ref{sec:eval-research} recovers such a failure by rewriting the skill, so a broken translation costs attempts rather than the capability, though cross-family reuse remains unmeasured and is the natural next step.\looseness=-1

\noindent\textbf{Where determinism breaks.}
Astropy is the one category reuse does not replay, and the two astropy tasks account for the whole gap, since a warm-start run solves two of four, three of eight, and four of ten of them yet none on the first attempt, against 5 of 6, 12 of 12, and 13 of 15 on sympy.
Solving on the second attempt and never the first points at the environment rather than the model, and our hand-classified failures agree, since astropy's \texttt{conftest} is a known fault of the SWE-bench image, like the sphinx Jinja drift.
The first attempt is spent discovering that the environment, not the patch, failed, after which the capability applies cleanly, which bounds what a benchmark image can say about determinism and is where the research loop matters most.\looseness=-1

\section{Related Work}
\label{sec:related}

Work on agentic systems approaches our problem from four directions.
Multi-agent frameworks decide which agent acts at each step, memory systems give an agent a record of what it has seen, tool-selection methods reduce what invoking a tool costs, and a smaller body lets an agent construct capabilities of its own.
We take each in turn, closing with what \phronix{} does differently, which concerns where a harness keeps what it learned and whether that knowledge can be executed.\looseness=-1

\subsection{Multi-Agent Frameworks and Agent Generation}
Work in the first direction composes several agents to attack a task, where AutoAgents~\cite{chen_autoagents_2024} generates a team of role-specialised agents, Tressoir~\cite{latyr_ngom_tressoir_2026} expresses a multi-agent system as an interpretable blueprint editable online, offline, or with a human in the loop, and VibeCodeHPC~\cite{hayashi_vibecodehpc_2026} assigns manager, engineer, and programmer roles to tune HPC code on a supercomputer.
Composition decides which agent acts next, while the capabilities those agents call remain whatever their designer provided.
\phronix{} is orthogonal to this division of labour and changes the layer beneath, since every agent there still solves its step with given tools rather than built ones.\looseness=-1

\subsection{Agent Memory and Reflective Adaptation}
Work in the second direction gives an agent a memory of its past, where semantic and episodic memory~\cite{hassell_learning_2026} let it reflect on supervision received and adapt later behaviour, and FORGE~\cite{bogdanov_forge_2026} evolves that memory across a population by broadcasting what individuals learn, with no weight updates.
ExpeL~\cite{zhao2024expel} similarly learns from prior task experience by extracting natural-language insights and retrieving successful trajectories as in-context demonstrations.
These approaches accumulate a record read back into context, so the agent is reminded of what happened and must still reconstruct how to act on it.
\phronix{} accumulates executable capability instead, since an \ac{MCP} tool paired with its skills runs the solution rather than describing it, which makes a recurring task solvable on the first attempt.\looseness=-1

\subsection{Tool Selection and Tool-Use Efficiency}
Work in the third direction reduces what tool use costs, where AutoTool~\cite{jia_autotool_2026} learns the sequential patterns of past trajectories and selects the next tool and its arguments by traversing a graph rather than asking the model again, cutting inference cost by up to 30\%.
Such work assumes a tool set that already covers the task and optimises the choice within it.
\phronix{} addresses where that tool set comes from, and the two compose naturally, since a store growing by tens of tools per task is exactly where cheaper selection pays.\looseness=-1

\subsection{Self-Evolving Capability Construction}
Closest to our work, Alita~\cite{qiu_alita_2025} equips a generalist agent with minimal predefinition and lets it construct, refine, and reuse \ac{MCP} servers generated from open source, reaching 75.15\% pass@1 on GAIA, while tool-making frameworks such as LATM~\cite{cai2024large} and CREATOR~\cite{qian2023creator} let LLMs construct reusable tools.
\phronix{} differs from these approaches by learning capabilities from complete harness executions, separating deterministic behavior into executable tools from contextual judgment retained as skills, and persisting the resulting capabilities in a layered cross-task store.
Domain agents such as IOAgent~\cite{egersdoerfer_ioagent_2025} reach expert-level I/O diagnosis by retrieving curated domain knowledge into context.
\phronix{} differs from a retrieval-based domain agent such as IOAgent in what it leaves behind, since diagnosing a trace with retrieved knowledge ends with the answer whereas instrumenting an application with built capabilities ends with tools the next workflow reuses.\looseness=-1

\section{Conclusions}
\label{sec:conclusions}

We presented \phronix{}, a learning paradigm that makes building capabilities the first-class citizen of the harness and runs it as a predictor--validator--corrector loop, where a capability of tool and skills predicts, a validator built from tools alone rules on what it produced, and those findings correct the capability rather than working around it.
An agent therefore learns from its own execution with no user in the loop, and the improvement it shows between attempts decides whether a task continues.
Such capabilities solve a recurring task on the first attempt in 93--95\% of cases across every model, turn an agent's own execution into 7.6--30.8 durable capabilities per task without instruction, and solve a task for 0.38--0.40M tokens, three to four times less than a single baseline attempt.
On the 1000 Genome Pegasus workflow, \phronix{} instruments a production scientific application for I/O performance analysis correctly on all 11 checks against 6 for the same harness without it, and it lifts gpt-5.4 and gpt-5.5 from 29\% and 27\% of SWE-bench tasks solved to 91\% and 87\%, level with the costliest model we measured, so a facility pays for its routine engineering once rather than on every repetition.\looseness=-1

\section*{Acknowledgment}
This work was performed under the auspices of the U.S. Department of Energy by Lawrence Livermore National Laboratory under Contract DE-AC52-07NA27344 (LLNL-CONF-2024996). This material is based upon work supported by LLNL LDRD 25-ERD-042.

\balance

\bibliographystyle{IEEEtranDOI}
\bibliography{references}

\end{document}